\documentclass{osa-article}

\journal{osajournal}
\usepackage{graphicx}
\usepackage{amsmath,amssymb} 
\usepackage{subcaption}
\newcommand\Illum[1]{E_{#1}}
\newcommand\Reflect[1]{r_{#1}}
\newcommand\Surf[1]{\mathcal{S}_{#1}}
\newcommand\Point[1]{P_{#1}}

\newcommand\Normal[1]{\vec{N_{#1}}}
\newcommand\Dsurf[1]{dP_{#1}}

\newcommand\rh[2]{\rho^{#1}_{#2}}

\begin{document}

\title{Deep Spectral Reflectance and Illuminant Estimation from Self-Interreflections}

\author{Rada Deeb,\authormark{1,*} Joost Van de Weijer,\authormark{2,*} Damien Muselet,\authormark{1} Mathieu Hebert, \authormark{1} and Alain Tremeau\authormark{1}}

\address{\authormark{1}University of Lyon, UJM-Saint-Etienne, CNRS, Institut d'Optique Graduate School, Laboratoire Hubert Curien UMR 5516, F-42023, SAINT-ETIENNE, France\\
\authormark{2}Computer Vision Center, Universitat Autonoma de Barcelona, Spain}

\email{\authormark{*}rada.deeb@univ-st-etienne.fr} 




\begin{abstract}
In this work, we propose a CNN-based approach to estimate the spectral reflectance of a surface and the spectral power distribution of the light from a single RGB image of a V-shaped surface. Interreflections happening in a concave surface lead to gradients of RGB values over its area. These gradients carry a lot of information concerning the physical properties of the surface and the illuminant. Our network is trained with only simulated data constructed using a physics-based interreflection model. Coupling interreflection effects with deep learning helps to retrieve the spectral reflectance under an unknown light and to estimate the spectral power distribution of this light as well. In addition, it is more robust to the presence of image noise than the classical approaches. Our results show that the proposed approach outperforms the state of the art learning-based approaches on simulated data. In addition, it gives better results on real data compared to other interreflection-based approaches.
\end{abstract}


\section{Introduction}
\begin{figure}[h]
	\centering
		\includegraphics[width=0.7\textwidth]{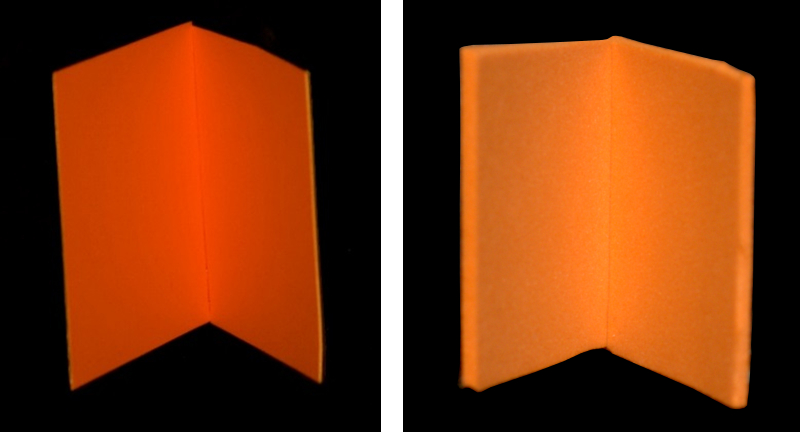}
	\caption{Examples of interreflections. Note how the colors are changing when moving towards the fold.}
	\label{fig:interreflection}
\end{figure}

When one observes an isolated flat paper under perfect diffuse light, no color variation appears and each elementary surface provides the same information about the light and the surface reflectance of the paper.
Most of the time, this information is not rich enough to deduce anything about the light or the surface. For example, a paper appearing blue can be a blue paper under white light or a white paper under blue light. If one decides to fold the paper, the beams coming from the light source will bounce between the elementary surfaces thereby creating color variations across the paper (see Fig.\ref{fig:interreflection} and \ref{fig:resReal}). This phenomenon is called interreflections and is well modeled by physics-based equations~\cite{Nayar91,Seitz05,Deeb2017}. These models clearly explain that two elementary surfaces appear with different colors because they received different spectral lights. So creating interreflections in a scene can be compared to observing the surface under a wide range of different lights. Thus each elementary surface provides different and complementary information about the original light and the surface reflectance. This is why interreflections can be considered as extra information to extract physical properties of the observed scene.

The problem of interreflection estimation has typically been addressed with physics-based approaches \cite{Nayar91,Nayar92,Seitz05,Deeb2017}. These methods are based on image formation models which consider intrinsic properties such as material reflectances, camera sensitivity, illuminant colors, and scene geometry. However, finding the inverse function of these image formation models is known to be very difficult. It is only after imposing additional assumptions that these problems can be solved. These assumptions include Lambertian surfaces, a single known illuminant in a scene, a known geometry,  etc. They also include more technical assumptions regarding the absence of noise; the reason for which these algorithms are often evaluated on high quality laboratory images or even only on synthetic data. The advantage of these methods is that when the assumptions hold they obtain exact solutions. However, in more realistic situations - as our experiments will show – the assumptions do not hold; and for example the presence of noise, a change in geometry, and non spatially homogeneous light can greatly influence the quality of the estimation of the scene intrinsic parameters.

In recent years, deep learning algorithms have shown an impressive ability to learn non-linear functions for problems were abundant data is available. Even though these networks often consist of millions of parameters they do not tend to overfit quickly. The advantage of using deep networks for deriving an inverse function is that we only require the image formation equations which are typically simpler than their inverse. The field of computer graphics studies realistic models for image formation \cite{cohen85,jensen01,pharr16}. With these equations we can generate training data consisting of generated scenes and known input intrinsic parameters. In addition, we can add noise to the process which more realistically models the world. After generation of the data we can use deep networks to find the inverse function and return the intrinsic parameters of the scene given an input image. In conclusion, using deep networks to find inverse functions of the image formation model has two advantages: 1. for many realistic image formation settings there are currently no inverse functions known, however when the forward image formation equations are known we can generate the data to train a deep network, and 2. we can easily consider more realistic scenarios, e.g. by adding noise in the image formation model, which make the system robust to situations which might be hard to model by physics-based methods. 

In this paper, we investigate the application of deep network to find the inverse function for interreflections.  We first train a network on simulated data built using an infinite-bounce physics-based interreflection model \cite{Deeb2017,Deeb2018}. Next, we loosen the assumption of a known illuminant and leave this task also to the network. For this case no inverse solution is currently available. We show that: 

\begin{enumerate}
\item Even though our datasets are constituted of only physics-based simulated data, our proposed method significantly outperforms the physics based methods in the presence of realistic settings with noise.
\item Deep networks can be applied to inverse problems for which no solution exists, i.e. estimation of interreflection with unknown illuminant. In experiments, we obtain better results than approaches that require knowledge of the illuminant spectral distribution.
\item The proposed networks can provide an accurate estimate of the material reflectance from a single interreflection compared to kernel-based learning approaches. The scene illuminant is estimated with good accuracy too. This has potential applications for color measurements for online shopping, fruit quality assessment from mobiles, etc.
\end{enumerate}

The paper is organized as follows. In Section~\ref{sec:Related Work} we discuss the related work. In Section~\ref{sec:datasets} we describe the method used to create the dataset. In Section~\ref{sec:network} we describe the deep network architecture we propose for reflectance and light estimation. In Section~\ref{sec:experiments} we show the experimental results, and finally in Section~\ref{sec:conclusions} we conclude.

\section{Related Work}\label{sec:Related Work}

\subsection{Spectral Reflectance Estimation}

State of the art in spectral reflectance estimation of surfaces from RGB data can be divided into two categories: direct methods as called in \cite{zhao07,ribes08}, also named observational-model based methods in \cite{heikkinen16}, and indirect methods as called in \cite{zhao07,ribes08}, also named learning-based methods in \cite{heikkinen16}. Spectral responses of sensors and spectral power distribution (SPD) of illuminants are considered to be known when direct methods are used. A common approach of these methods is to combine trichromatic imaging with multiple light sources \cite{park07,heikkinen08,chi10,fu16,han14,jiang12,khan13}. Park et al. \cite{park07} obtained spectral information by using RGB camera with a cluster of light sources with different spectra. They model the spectral reflectance of surfaces with a set of basis functions in order to reduce the dimensionality of the spectral space. Later, Jiang et al. \cite{jiang12} proposed to use commonly available lighting conditions, such as daylight at different times of a day, camera flash, ambient, fluorescent and tungsten lights. More recently, Khan et al. \cite{khan13} proposed the use of a portable
screen-based lighting setup in order to estimate the spectral reflectance  of the considered surface. The portable screen was used to give three lightings with green, red and blue colors and the spectral reflectance of the surface is expressed by nine coordinates in a basis of nine spectra. These basis functions are obtained by eigendecomposition of spectral reflectances of 1257 Munsell color chips discarding the wavelengths corresponding to the sensor-illuminant null-space. 

On the other hand, learning-based approaches do not require prior knowledge of the spectral response of the sensors or the SPD of the lighting system. In addition, they can be used with a single light source \cite{heikkinen07,heikkinen08,heikkinen13,heikkinen16}. In~\cite{heikkinen16}, screen-based lighting is used to reconstruct spectral reflectances from multiple images. In this last paper, the used approaches are the ones introduced in~\cite{heikkinen13}, which are mostly based on the non-linear ridge regression and whose differences are in the choice of the used kernel (e.g. polynomial or Matérn) or of the link function (e.g. logit or Gaussian copula). The link functions are applied as a pre-processing step to the input reflectance vectors before learning and their inverse counterpart are applied to the output vectors to recover the reconstructed reflectance. Their main aim is to constrain the reconstructed spectra to be in the range [0,1] due to physical reasons.However, these last approaches depend on the quality of the training set and on the choice of the regression method. It has been shown in \cite{heikkinen16}, that when no high quality training set is available, using multiple light sources becomes important in order to improve the quality of the results.

Recently, some works aimed at recovering the spectral curves of all the pixels of a single RGB image~\cite{Arad2016,Nguyen2014,Aeschbacher_2017,Alvarez2017}. The idea of~\cite{Nguyen2014} is to model the mapping between camera RGB values and scene reflectance spectra with a radial basis function network.  Arad et al.~\cite{Arad2016} exploit the prior on the distribution of hyperspectral signatures in natural images in order to construct a sparse hyperspectral dictionnary. The projection of this dictionary into RGB provides a mapping between RGB atoms and hyperspectral atoms. Thus, given RGB values that can be decomposed into RGB atoms, their spectral reflectance is obtained by using the same combination of the corresponding hyperspectral atoms. The accuracy and efficiency of this approach was improved by Aeschbacher et al.~\cite{Aeschbacher_2017} who proposed to learn the dictionary projection between the training RGB and hyperspectral data and to extract anchor points from this projection. At test time, a simple nearest anchor search is run for each RGB triplet in order to reconstruct its spectral curve. Finaly, Alvarez et al. proposed a convolutional neural network architecture that learns an end-to-end  mapping  between pairs of input RGB images and their hyperspectral counterparts~\cite{Alvarez2017}. They adopt an adversarial framework-based generative model that takes into account the spatial contextual information present in RGB images for the spectral reconstruction process. For these last works, even if the aim is also to reconstruct spectral functions from RGB data, the objective appears to be different from the task addressed in this paper. Recovering a full resolution hyperspectral image is much more challenging and can not provide results as accurate as the ones provided by approaches that concentrate on a single surface observed under calibrated conditions. For example, the best results in terms of root mean square error (RMSE) reported in~\cite{Aeschbacher_2017} are more than $1.00$, while the RMSE provided by~\cite{Deeb2018} on real images is around $0.05$.

\subsection{Interreflections}
Interreflections refer to the phenomenon that each point in a concave surface reflects light towards each other point, and thus re-illuminates it to a more or less extent according to its reflective properties and the geometrical shape of the surface. The simplest case is a flat diffusing surface bent into two flat panels with an angle between them.

This phenomenon has been studied in the domain of computer vision, mainly in order to remove this effect from the images to be able to retrieve the shape of an object in an image (shape-from-shading methods) \cite{Nayar91,Nayar92,Seitz05,Funt93,Liao11,fu14}. Recently, interreflections have been exploited in the context of color camera calibration~\cite{Deeb_bmvc17}. Some approaches in the literature~\cite{Funt91,drew90,ho90} used interreflections as extra source of information in order to obtain the surface spectral reflectance and the light SPD. Only adjacent panels having different spectral reflectances are used while taking into consideration one bounce of interreflected light. Recently, Deeb et al.~\cite{Deeb2018} exploited infinite bounces of interreflections in order to help the estimation of the spectral reflectance of surfaces. Given the light SPD and the camera spectral responses, they formulate this estimation as an optimization problem. We use their physics-based interreflection model to create the training dataset but instead of solving the inverse problem explicitly, we apply deep learning to learn the inverse function. 

\subsection{Deep Learning for Physics Based Vision}
Deep learning approaches have also been applied to physics based vision, where most of the works try to solve the color constancy problem. For example, since the choice of the pooling between local and global estimates is not an easy task, Fourure et al.~\cite{Fourure2016} proposed a deep network that can choose between the different poolings. The output of this network was a global light color estimate. Shi et al.~\cite{Shi2016} rather proposed two interacting sub-networks that locally estimate the light color. Their idea is to create multiple hypotheses for each patch with the first sub-network and then use the second one to vote for the best hypotheses. Recently, Hu et al.~\cite{Hu2017} proposed a fully convolutional network to create a confidence map that selects (weights) the patches in the image which provide the best light color estimate. On the other hand, deep networks were also applied to intrinsic images decomposition. For example, Narihira et al.~\cite{Narihira2015} proposed to use a convolutional neural network to predict lightness difference between pixels learned from human judgment on real images. Shi et al.~\cite{Shi2017} extended this approach to non-Lambertian objects and proposed a CNN that is able to recover diffuse albedo, shading, and specular highlights from a single image of an object. Recently, Janner et al.~\cite{Janner2017} proposed the Rendered Intrinsic Network that contains two convolutional encoder-decoders: one to decompose the input image into reflectance, shape and lighting and another that is reconstructing the input for these resulted images. The advantage of this network is that it can learn from unlabeled data because it is using the unsupervised reconstruction error as a loss function.

\begin{figure*}[ht]
	\centering
	\begin{subfigure}[b]{0.3\textwidth}
		\includegraphics[width=\textwidth]{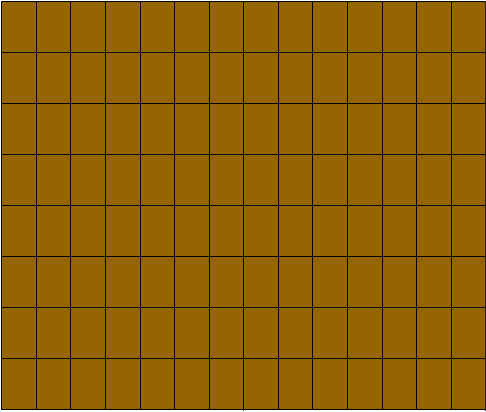}
		\caption{Flat brown surface under a white light.}
		\label{fig:metaFlat1}
	\end{subfigure} \hspace{1cm}
	\begin{subfigure}[b]{0.3\textwidth}
		\includegraphics[width=\textwidth]{./metamerismFlat.jpg}
		\caption{Flat white surface under a brown light.}
		\label{fig:metaFlat2}
	\end{subfigure}
	
	\begin{subfigure}[b]{0.4\textwidth}
		\includegraphics[width=\textwidth]{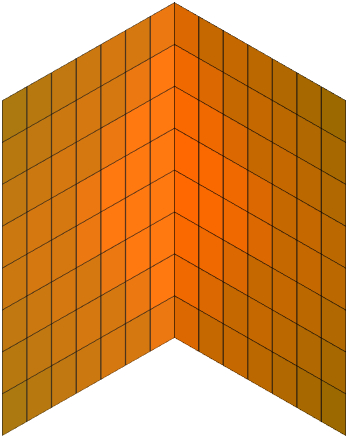}
		\caption{Folded brown surface under a white light.}
		\label{fig:metaFold1}
	\end{subfigure}
	\begin{subfigure}[b]{0.4\textwidth}
		\includegraphics[width=\textwidth]{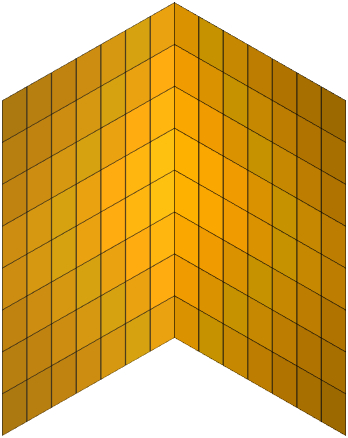}
		\caption{Folded white surface under a brown light.}
		\label{fig:metaFold2}
	\end{subfigure}
	\caption{Interreflections and metamerism: two Metameric flat surfaces show no metamerism when folded with an angle of $45^\circ$. The graphics are computed with the spectral infinite-bounce  interreflection model explained in Section \ref{sec:datasets}.   }
	\label{fig:metaEx}
\end{figure*}

\section{Motivation} \label{sec:motivation} 
Our work is based on the hypothesis that from interreflections only a neural network can learn to estimate both the spectral reflectance of the surface and the spectral power distribution of the light. In order to demonstrate the motivation behind this hypothesis, let us study how interreflections happen and what kin of information they hold. When a point on a concave surface receives a light ray from the light source, this ray might bounce several times before being able to exit the surface. The total energy carried with a light ray when it arrived to the camera sensor is the sum of the energies it carried with each bounce it encounters. In fact, with each bounce the energy carried with the light ray decreases. The sooner this energy vanishes the less bounces the light ray carrying it encounters.

It has been demonstrated  in~\cite{Deeb2017,Deeb2018,Nayar91} that the total irradiance at a given point of a concave surface is the sum of the irradiance directly received from the light source and the indirect irradiance coming from multiple reflections of the source on the other points constituting the surface. For a given wavelength, the irradiance received by a point $\Point{1}$ of a Lambertian surface $\Surf{}$ after a single bounce of the light source on the other points $\Point{i}$ $\in$ $\Surf{}$ with infinitesimal area $\Dsurf{i}$ and characterized by the reflectance $\Reflect{i}$, is defined as:

\begin{align} \label{eq:E1Int}
\Illum{1}(\Point{1}) = \int_{\Point{i} \in \Surf{}} \Reflect{i} \frac{\Illum{0}}{\pi} K(\Point{i}, \Point{1})\Dsurf{i}, 
\end{align}
where $\Illum{0}$ is the irradiance received form direct light source and $K(\Point{i},\Point{j})$ is the geometrical kernel $K$ defined for every pair of points, $\Point{i}$ and $\Point{j}$ as:
\begin{equation}  \label{eq:kernel}
K(\Point{i},\Point{j}) =  \frac{(\Normal{i}\cdot\vec{\Point{i}\Point{j}})(\Normal{j}\cdot\vec{\Point{j}\Point{i}})V(\Point{i},\Point{j}) }{(\Delta_{ij})^4}.
\end{equation}
The vectors $\Normal{i}$ and $\Normal{j}$ are the surface normals at $\Point{i}$ and $\Point{j}$, $\Delta_{ij}$ is the Euclidean distance between the two points, and $V(\Point{i},\Point{j})$ is a visibility term  which takes $1$ if the areas around these points can see each other and 0 otherwise.

Similarly, after two bounces of light ray the irradiance can be written as:

\begin{equation}\label{eq:E2Int}
\Illum{2}(\Point{i}) = \int_{\Point{j} \in \Surf{}}\int_{\Point{j'} \in \Surf{}} \Reflect{j} \Reflect{j'} \frac{\Illum{0}}{\pi^2} K(\Point{i}, \Point{j})K(\Point{j}, \Point{j'})\Dsurf{j}\Dsurf{j'},
\end{equation}

One can observe from these equations that,  in contrast to highlights, the spectral radiation of a light ray is changed with each bounce depending on the spectral reflectance of the point it hits. Thus, each bounce can be considered as a new light source with different spectral properties. In addition, the number of bounces exchanged between a pair of points, $\Point{i}$ and $\Point{j}$ is related to their relative geometrical relation defined by the term $K(\Point{i}, \Point{j})$. Close face to face points have big geometrical kernel values, thus they would exchange a high number of rays, while far points will exchange much less, having a small value in the corresponding geometrical kernel. However, even for two close points with a high kernel value, the energy carried with a light ray reflecting from $\Point{j}$ toward $\Point{i}$ will be low at a given wavelength if $\Reflect{j}$ is low at this wavelength. The number of bounces and the energy carried with each bounce are the  cause of the color gradients over the concave surface. 

Let us take into consideration a special case of interreflection: self-interreflections happening over a Lambertian surface which is the case we are studying in this paper. In this case, the surface has the same spectral reflectance, $\Reflect{}$ all over its area. Then, previous equations can be written as: 

\begin{align} \label{eq:E1IntS}
\Illum{1}(\Point{1}) = \int_{\Point{i} \in \Surf{}} \Reflect{} \frac{\Illum{0}}{\pi} K(\Point{i}, \Point{1})\Dsurf{i}, 
\end{align}

\begin{equation}
\Illum{2}(\Point{i}) = \int_{\Point{j} \in \Surf{}}\int_{\Point{j'} \in \Surf{}}  \Reflect{}^2 \frac{\Illum{0}}{\pi^2} K(\Point{i}, \Point{j})K(\Point{j}, \Point{j'})\Dsurf{j}\Dsurf{j'},
\end{equation}

From these equations, one can observe that the RGB values of an area with interreflections have different relations with the surface spectral reflectance and the light SPD. With each bounce the reflectance is risen to a higher power, whereas the relation with light SPD is kept linear. 


One of the advantages of the asymmetry between the surface reflectance and light SPD is that it allows us to distinguish materials which might appear as metamers in the absence of interreflections. As an example consider Figure~\ref{fig:metaEx}  where we show an example of metamerism: a flat brown surface with a white illuminant is indistinguishable from a white surface with a brown illuminant. However, when we fold the surface and consider inter-reflections the observed surfaces do significantly differ. We therefore argue that two surfaces which show metamerism when they are flat, have almost no chance to show the same color gradients when they are folded into a concave shape. For this reason, interreflections can be seen as an important sources of information to get both surface reflectance and light SPD while avoiding the case of metamerism. In this paper, we investigate the usage of deep networks to extract intrinsic scene information from interreflections.

\section{Datasets} \label{sec:datasets}
Simulated datasets built using an infinite-bounce interreflection model for Lambertian surfaces are considered. In this section, we first present the used model, then the proposed datasets and the data augmentation performed on them.
The used model in the one proposed by Deeb \textit{et al.} \cite{Deeb2018}. In this work, we adopt this model to build simulated datasets to be used to train the network.

\subsection{The interreflection Model} \label{sec:model}

Starting from Equations (\ref{eq:E1Int}) and (\ref{eq:E2Int}), a discrete version of the model was proposed by Nayar \textit{et al.} \cite{Nayar91}, and can be obtained by sampling the surface into a finite number $m$ of small facets, where each facet is assumed uniformly illuminated, flat and uniform in reflectance. The area of a facet centered on a point $\Point{i}$ is denoted $S_i$ and $K_{ij}$ is the geometrical kernel between $\Point{i}$ and $\Point{j}$. The irradiance in a facet centered on $\Point{1}$, after one bounce can be re-written as:

\begin{equation} \label{eq:Esum1}
\Illum{1}(\Point{1}) = \sum_{i=1}^{m} \Reflect{i} \frac{\Illum{0}}{\pi} K_{i1} S_i,
\end{equation}

Similarly, if the light rays reflect twice on every pair of points $\Point{i}$ and $\Point{j}$ with respective reflectances $\Reflect{i}$ and $\Reflect{j}$, the irradiance received by a point $\Point{1}$ is:
\begin{equation} \label{eq:Esum2}
\Illum{2}(\Point{1}) = \sum_{j=1}^{m}\sum_{i=1}^{m} \Reflect{j} \Reflect{i} \frac{\Illum{0}}{\pi^2} K_{ij} K_{j1} S_j S_{i}.
\end{equation}

These equations can be written in a matrix form: 
\begin{equation} 
\mathbf{E_1} =   \mathbf{K}  \mathbf{R}  \mathbf{\Illum{0}},
\end{equation} 
and
\begin{equation} 
\mathbf{E_2} =   (\mathbf{K}  \mathbf{R})^2  \mathbf{\Illum{0}},
\end{equation} 
where $\mathbf{K}$ is a square matrix, symmetric when all the facets are of equal size, with zeros on the diagonal due to the fact that rays cannot transit to a facet from themselves:

\begin{equation} \label{eq:kernel_matrix2}
\mathbf{K} = \frac{1}{\pi} \begin{bmatrix}
0 & K_{12} S_{2} &. & . & K_{1m} S_{m}\\
K_{21} S_{1} & 0 & . & . & K_{2m} S_{m} \\
. &. & 0 &. &. \\
K_{m1} S_{1} & .& . & . & 0
\end{bmatrix} ,
\end{equation}
$\mathbf{R}$ is a diagonal matrix grouping all spectral reflectances  of the different facets for the considered wavelength:

\begin{equation}  \label{eq:reflect_matrix}
\mathbf{R} = \begin{bmatrix}
\Reflect{1} &0 &...... & 0\\
0 & \Reflect{2} & .... & 0 \\
. & .& . & .... \\
0 & 0& .. & \Reflect{m} \\
\end{bmatrix},
\end{equation}
and $\mathbf{\Illum{0}}$ is a vector of dimension $m$ representing the direct irradiance received by the $m$ facets.

Based on this matrix form, the cumulated irradiance values on each facet after $n$ bounces of light, grouped into an irradiance vector of size $m$, can be written as:

\begin{equation}  \label{eq:sumIllum}
\mathbf{E_{0\rightarrow n}}= \sum_{b=0}^{n} \mathbf{E_b} =  \sum_{b=0}^{n}  ( \mathbf{K}  \mathbf{R})^b  \mathbf{\Illum{0}}.
\end{equation} 

This sum corresponds to a geometric series. In case of non fluorescent surfaces, when $n$ tends to infinity, this series converge to:

\begin{equation} \label{eq:illum}
\mathbf{E}  = ( \mathbf{I} -  \mathbf{K}  \mathbf{R})^{-1}   \mathbf{\Illum{0}}.
\end{equation}

This is a general expression of irradiance for a single wavelength after infinite bounces of light for Lambertian surfaces. A more handy form can be obtained from this equation by writing the radiance reflected towards the camera in terms of direct irradiance ($\mathbf{L} = \frac{1}{\pi}  \mathbf{R} \mathbf{E}$):

\begin{equation} \label{eq:rad}
\mathbf{L} = \frac{1}{\pi} (  \mathbf{R}^{-1} -   \mathbf{K})^{-1}   \mathbf{\Illum{0}}.
\end{equation}

Since we are working with RGB values captured by camera sensors, we need to integrate this radiance value over the wavelength range of the sensor sensitivities. So we need to use an extended version of this equation (defined for a single wavelength), in order to take into consideration all pixels and all wavelengths simultaneously. This version has been proposed in~\cite{Deeb2017} as follow:
\begin{equation} \label{eq:rad_ext}   
\mathbf{L_{ext}} = \frac{1}{\pi} (  \mathbf{R_{ext}}^{-1} -   \mathbf{K_{ext}})^{-1}   \mathbf{\Illum{0_{ext}}},
\end{equation}
where
$\mathbf{\Illum{0}}_{ext}$ is a vector of length $mq$ which is obtained by concatenating the wavelength-specific $\mathbf{\Illum{0}}$ vectors.
$\mathbf{R}$ is extended to another square diagonal matrix $\mathbf{R}_{ext}$ of size $mq \times mq$, where $q$ is the number of wavelengths. Renaming $\mathbf{R}$ defined for the wavelength $\lambda_i$ as $\mathbf{R}_{\lambda_i}$, $\mathbf{R}_{ext}$  is expressed as: 

\begin{equation}
\mathbf{R}_{ext} = \begin{bmatrix}
\mathbf{R}_{\lambda_1} &0 &...... & 0\\
. & \mathbf{R}_{\lambda_2}& ..... & . \\
. &  .... & . & 0 \\
0 & .& ...... & \mathbf{R}_{\lambda_q}\\
\end{bmatrix}.
\end{equation}

Likewise, $\mathbf{K}$ is extended to the square matrix $\mathbf{K}_{ext}$ of size $mq \times mq$:

\begin{equation}  \label{eq:K_ext}
\mathbf{K}_{ext} = \begin{bmatrix}
\mathbf{K} &0 &...... & . & . & . &.& 0\\
. & \mathbf{K}& . & .... & . & . &.& . \\
. & . & . & .... & . & . &.& . \\
0 & 0& .. & .  & . & . &...... & \mathbf{K}\\
\end{bmatrix}.
\end{equation}

Let consider $s$ sensors ($s=3$ for classical RGB cameras) whose spectral sensitivities are inserted in the $ms \times mq$ matrix $\mathbf{C}_{ext}$ as follow: 
\begin{equation}
\mathbf{C}_{ext} = \begin{bmatrix}
\mathbf{C^1}_{\lambda_1} & \mathbf{C^1}_{\lambda_2} &...... & . & . & . &.& \mathbf{C^1}_{\lambda_q}\\
\mathbf{C^2}_{\lambda_1} & \mathbf{C^2}_{\lambda_2} &...... & . & . & . &.& \mathbf{C^2}_{\lambda_q} \\
. & .& . & .  & . & . &. & .\\
\mathbf{C^s}_{\lambda_1} & \mathbf{C^s}_{\lambda_2} &...... & . & . & . &.& \mathbf{C^s}_{\lambda_q} \\
\end{bmatrix}.
\end{equation}
where the $m \times m$ matrix $\mathbf{C^i}_\lambda$ is associated to the sensor $i$:
\begin{equation}
\mathbf{C^i}_\lambda = c^i_{\lambda} I_m,
\end{equation}
where $I_m$ is the m-dimensional identity matrix.

Thus, the camera sensor responses can be obtained as a $ms$-dimensional vector:
\begin{equation}
\boldsymbol{\rho}_{ext} = \begin{bmatrix}
\rh{1}{1} \quad . .  \quad \rh{1}{m} \quad \rh{2}{1} \quad ..  \quad \rh{2}{m} \quad  ... \quad \rh{s}{1} \quad .. \quad \rh{s}{m}
\end{bmatrix}^T,
\end{equation}
thanks to the equation:
\begin{equation}   \label{eq:spectral}
\boldsymbol{\rho}_{ext} =  \frac{1}{\pi} \mathbf{C}_{ext}  ( \mathbf{R}^{-1}_{ext} -  \mathbf{K}_{ext})^{-1}  \mathbf{\Illum{0 ext}}.
\end{equation}

Since we consider in this paper only self-interreflections, we assume that all the facets of a considered surface have the same reflectance. Consequently, we can exploit the eigendecomposition of $\mathbf{K}_{ext}$ proposed by Deeb et al. in~\cite{Deeb2018} in order to speed up the evaluation of the camera sensor responses from equation~(\ref{eq:spectral}).
  
\subsection{Self-Interreflection Dataset Construction}

To train our network, we created datasets of simulated images of size $10 \times 10$ corresponding to one side of a folded V-shaped surface for each of the $1269$ Munsell patches. A dataset is built for a specific camera with known spectral response functions and a specific angle. Datasets are built taking into consideration multiple light sources. However, they can be built using a single light source also.

Each image in the dataset corresponds to a V-shaped configuration of a surface with a homogeneous spectral reflectance. In order to do this, geometrical kernel values are obtained using Monte Carlo integration after choosing the angle, the sizes of the planar surfaces, and the discretization size as suggested in \cite{Deeb2017}. We consider a single collimated light source, parallel to the bisecting plane of the two panels, illuminating the V-cavities frontally. As a consequence the irradiance received at each facet of the V-cavities from direct light is considered constant:

\begin{equation} \label{eq:E0eq}
\forall i,j:  \Illum{0}(\Point{i}) = \Illum{0}(\Point{j}) .
\end{equation}

For a surface with a known spectral reflectance, and after choosing the camera spectral response functions and the spectral power distribution of illuminant, RGB values can be obtained using the previously explained interreflection model.

The images are then pre-processed by applying mean subtraction and normalization. Data augmentation is performed at batch level by adding different levels of noise to each image.  We adopt two types of noise to be added to images at batch level: Poisson noise is added to each image, and a Gaussian noise of one of 5 different variances might be added to some images based on a random decision.

%

\section{Network Structure \& Loss Functions } \label{sec:network}

\begin{figure*}[t]
\centering
\includegraphics[height=4.5cm]{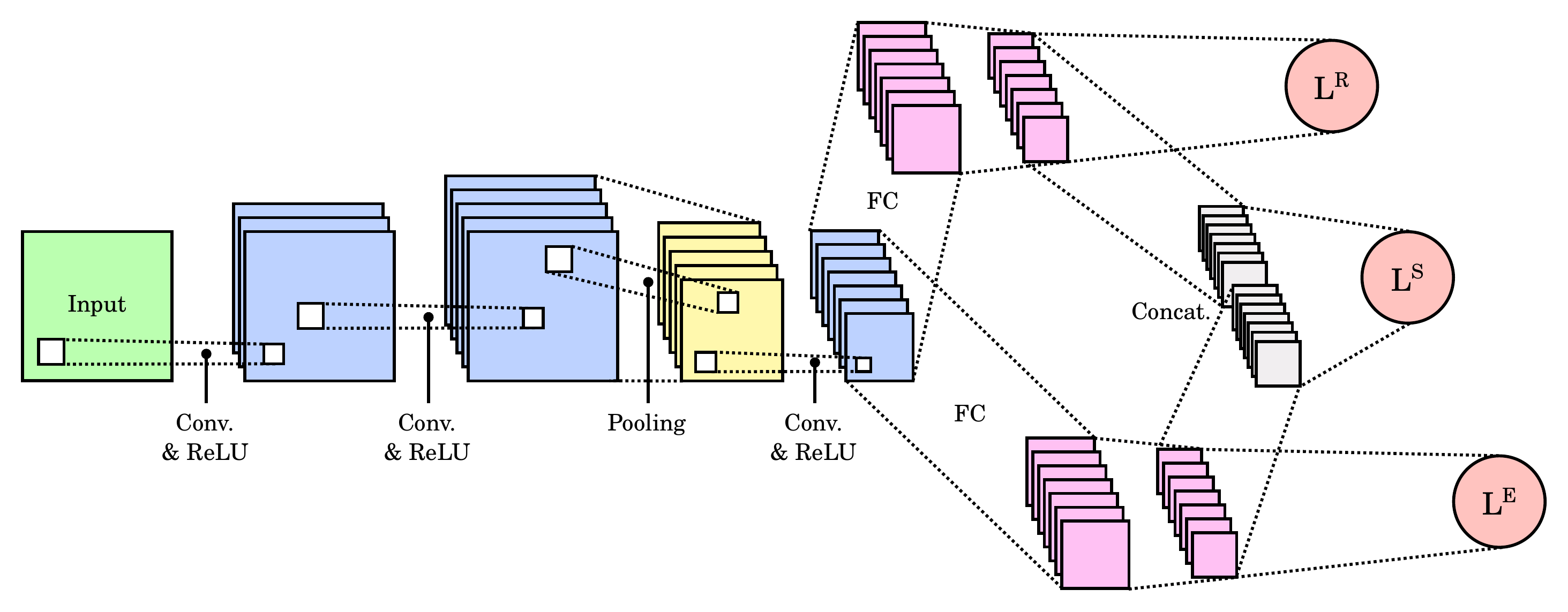}
\caption{Diagram of the proposed network for surface reflectance and illuminant estimation from interreflections.}
\label{fig:netStruct}
\end{figure*}

For our regression problem, we use the convolutional neural network structure presented in Figure \ref{fig:netStruct}, in order to obtain both the spectral reflectance of the surface and the spectral power distribution of the lighting from raw image RGB values. The network architecture starts with a shared part which consists of three convolutional layers; the first and second convolutional layers are of size $5 \times 5$ with padding of $2$. The size of the input starts to get smaller with the pooling layer of size $2 \times 2$ and then with the latest convolutional layer of size $3 \times 3$ and a stride of 3.  Subsequently, the network is split into two branches; one for the estimation of the surface reflectance and one for the spectral power distribution. These specialized branches consist  of two fully connected layers each. 

Let us denote by $\mathbf{R}_\lambda$, $\widehat{\mathbf{R}}_\lambda$, the ground truth spectral reflectance vector and the estimated one, respectively, and by  $\mathbf{E_0}_\lambda$, $\widehat{\mathbf{E_0}}_\lambda$, the ground truth SPD and the estimated one respectively. Each of these vectors contains the values of all the considered wavelengths. For this reason we put the subscript $\lambda$ to distinguish these vectors from the ones introduced previously. 

Three loss functions are used to obtain accurate spectral reflectance and SPD. The first loss function, $L^R(\mathbf{R}_\lambda)$ is regarding the error on the spectral reflectance defined as:
\begin{equation}
\L^R(\mathbf{R}_\lambda) = \frac{1}{2} (\mathbf{R}_\lambda -\widehat{\mathbf{R}}_\lambda )^2.
\end{equation}
The second loss function is related to the error in the SPD of illuminant and is defined as:
\begin{equation}
\L^E(\mathbf{E_0}_\lambda) = \frac{1}{2} (\mathbf{E_0}_\lambda -\widehat{\mathbf{E_0}}_\lambda )^2.
\end{equation}

In principle the network can be trained with only the $L^R$ and $L^E$ loss. However, in addition we introduce a consistency loss. We would like the multiplication of the estimated reflectance and illuminant to be close to the real color signal  $\mathbf{S}_\lambda=\mathbf{R}_\lambda \odot \mathbf{E_0}_\lambda$, where $\odot$ is the component-size multiplication operator so that $\mathbf{S}_\lambda$ is also a spectral signal. This loss enforces consistency between the separately estimated surface reflectance and light SPD. We use the CIE 1931 XYZ color matching functions to obtain an estimation which minimizes a perceptually relevant error because it gives more importance to wavelengths with high visual response~\cite{heikkinen13}. This consistency loss is defined as:   
\begin{align}
\L^S(\mathbf{S}_\lambda) = \frac{1}{2} (\mathbf{\bar{X}} \odot \mathbf{R}_\lambda \odot \mathbf{E_0}_\lambda -\mathbf{\bar{X}} \odot \widehat{\mathbf{R}}_\lambda \odot \widehat{\mathbf{E_0}}_\lambda)^2   \nonumber \\
+ \frac{1}{2} (\mathbf{\bar{Y}} \odot \mathbf{R}_\lambda \odot \mathbf{E_0}_\lambda -\mathbf{\bar{Y}} \odot \widehat{\mathbf{R}}_\lambda \odot \widehat{\mathbf{E_0}}_\lambda)^2 \nonumber \\
+ \frac{1}{2} (\mathbf{\bar{Z}}\odot \mathbf{R}_\lambda \odot \mathbf{E_0}_\lambda -\mathbf{\bar{Z}} \odot \widehat{\mathbf{R}}_\lambda \odot \widehat{\mathbf{E_0}}_\lambda )^2,
\end{align}
where $\mathbf{\bar{X}}$, $\mathbf{\bar{Y}}$ and $\mathbf{\bar{Z}}$ are the CIE 1931 XYZ color matching functions. 

The network is initialized using Xavier~\cite{glorot2010understanding}. The learning rate is set to $10^{-4}$ and is reduced every 20 epochs by a factor 10. The momentum is set to $0.9$ and the batch size to $50$.

The choice of network structure was done experimentally. The idea was to use a simple network that works well for our regression problem. We found that batch normalization cannot be used in these sort of regression problems where the exact intensity of the output signal is crucial for the quality of the estimation. Moreover, as the output spectral values for both surface and light are in the range $[0,1]$, one may think to add a sigmoid layer to accelerate the convergence. However, putting a sigmoid layer played the opposite role in our case leading to a  non convergence of the network. This is probably due to the fact that our network is relatively shallow, so a sigmoid is being an obstacle in the learning process.

In order to verify the importance of the third loss function,  $L^S$, we performed an ablation study. We trained the network twice under the same settings and on the exact same dataset, the only difference is that we deactivated the consistency loss $L^S$ for one training and activated it for the other.  Table \ref{table:consLoss} shows the percentage of enhancement in RMSE, PD and DE00 errors when the consistency loss is activated compared to when it is deactivated.

\setlength{\tabcolsep}{2pt}
\begin{table}[ht]
	\begin{center}
		\caption{Error enhancement when using the consistency loss}
		\label{table:consLoss}
		\begin{tabular}{lcccc}
			\hline\noalign{\smallskip}
			 & RMSE  &   PD & DE00  \\
			\noalign{\smallskip}
			\hline
			\noalign{\smallskip}
			Error enhancement (\%) & 7.7 & 11.6 & 14,8  \\
			
			\hline
		\end{tabular}
	\end{center}
\end{table}
\setlength{\tabcolsep}{1.4pt}

\section{Experiments \& Results}\label{sec:experiments}

In this section, our results on both simulated images and real camera outputs are presented. We consider a V-shaped configuration consisting of two planar square surfaces with an angle of $45^\circ$ between them (see Figure (\ref{fig:resReal})). The dataset and the test images contain only the area of one side of the planar surface discretized into $10 \times 10$ facets. 

Different metrics are used in order to evaluate our approach and to compare our results with those obtained using other spectral reflectance estimation approaches. The root mean square error (RMSE), and the Pearson distance (PD) are used to show the accuracy of spectral estimation. The performance is also evaluated in terms of color distance. We use CIEDE00 distance \cite{ohta06} with CIE D65 lighting and a $2^\circ$ viewing angle. This distance is computed between the color signal obtained with the ground truth spectral reflectance and the one obtained with the estimated spectral reflectance.   

\subsection{Simulated Data}
  For simulated data tests, the datasets are built using the CIE 1931 XYZ color matching functions as sensor spectral responses. This is done in order to compare our results to the ones in \cite{heikkinen13} under the exact same configurations. Wavelengths are taken between $400 nm$ a $700 nm$ with a $5 nm$ step, giving a total of $61$ wavelengths. Two datasets are considered for this case, the first is built using only the CIE D65 SPD. The network structure in this case is adapted so that the layers corresponding to the SPD of light are deactivated leading to only two loss functions $L^R$ and $L^S$. In addition, both $\mathbf{E_0}$ and $\widehat{\mathbf{E_0}}$ are replaced by the SPD of CIE D65 in the calculation of $L^S$. Following the same process as in \cite{heikkinen13}, the network is trained using $90 \%$ of Munsell patches and tested on the remaining $10 \%$.  The second dataset is built using $23$ different SPDs corresponding to illuminants on the Planckian locus with color temperatures ranging from $4000 K$ to $15000 K$ with steps of $500 K$  between them.  In this case, the network whose structure is shown in Figure (\ref{fig:netStruct}) is also trained using $90 \%$ of Munsell patches under various illuminants and tested on the remaining $10 \%$.

Our results on both datasets are shown in Table \ref{table:simResults}. They are compared to the best results regarding the average error in \cite{heikkinen13} obtained using a logit link function with different kernel models. The results are shown in terms of average, max and $95$th percentile RMSE and PD values. It can be observed from the results that when training the network with the dataset built using a single illuminant, which represents the setting of reflectance estimation with known illuminant, our approach outperforms the state of the art approaches. In addition, when training using the dataset built with different lightings, which is equal to the setting of unknown scene illuminant, our results are still better than the state of the art ones except for the maximum error values. However, in this case, the SPD of light is learned also with an average RMSE of $0.0154$. The average DE00 values are $0.6379$ and $0.7279$ for the single illuminant dataset and the multiple illuminant dataset respectively. It is worth mentioning that, in this case, our algorithm is able to reconstruct the unknown light SPD while for the other tested methods in Table \ref{table:simResults}, this SPD is a known input. 

\setlength{\tabcolsep}{2pt}
\begin{table*}
\begin{center}
\caption{Spectral error values for simulated data}
\label{table:simResults}
\begin{tabular}{lcccccc}
\hline\noalign{\smallskip}
Method & RMSE Avg. & RMSE Max. & RMSE 95th & PD Avg. & PD max. & PD 95th \\
\noalign{\smallskip}
\hline
\noalign{\smallskip}
Gaussian kernel & 0.0103 & 0.0508 & 0.0333 & 0.00104 & 0.0154 & 0.00462 \\
Matérn kernel & 0.0092 & 0.0558 & 0.0326 & 0.00088 & 0.0083 & 0.00420 \\
TPS kernel & 0.0092 & 0.0556 & 0.0332 & 0.00088 & 0.0083 & 0.00411 \\
Ours (1 light) & \textbf{0.0087} & \textbf{0.0460} & \textbf{0.0191} & \textbf{0.00081} & \textbf{0.0068} & 0.0041 \\
Ours (23 lights) & \textbf{0.0088} &   0.0771 & \textbf{0.0213} & 0.0012 & 0.0639 &  0.0041 \\

\hline
\end{tabular}
\end{center}
\end{table*}
\setlength{\tabcolsep}{1.4pt}

\subsection{Real Data}
For real data tests, a new simulated dataset is built using Munsell spectral reflectances, the Canon EOS 1000D spectral response functions and the 23 planckian illuminants. The network is trained with this dataset except the green and red patches in order to make sure that no overfitting is happening. The same camera is then used to take photos of V-shaped surfaces with an approximative angle of $45^\circ$ of six colored surfaces: a red Munsell paper, and five other textile pieces of different colors. The photos are taken under direct sunlight in the early afternoon. The area of the photo corresponding to one side of the V-shaped surface is selected manually to be then automatically discretized into $10 \times 10$ facets each represented by the mean RGB values over its area. In Table \ref{table:realResults} our results are compared to the state of the art ones on real images under a known illuminant \cite{park07,khan13,Deeb2017}. The results show that our approach outperforms the sate of the art ones in terms of spectral error even when a pre-calibration step using the XRite ColorChecker is performed to help these approaches. This pre-calibration is detailed in~\cite{Deeb2018} and is mandatory for the classical approaches \cite{park07,khan13} in order to get accurate reconstructions when used in non-calibrated configuration. It requires to add a color checker in the image, which is a strong constraint. Moreover, when compared to physics-based interreflection approach \cite{Deeb2018}, a significant enhancement in the spectral estimation is obtained when no pre-calibration step is used. The SPD of light is not known in our approach and is guessed with an average RMSE of $0.1$ in comparison to CIE D50 spectral power distribution.   

\setlength{\tabcolsep}{4pt}
\begin{table*}
\begin{center}
\caption{Results on real data}
\label{table:realResults}
\begin{tabular}{lcccccc}
\hline\noalign{\smallskip}
Method & Pre-calibration & Illuminant & RMSE  & PD  & DE00 \\
\noalign{\smallskip}
\hline
\noalign{\smallskip}
Park et al. \cite{park07} & Yes & Known & 0.060 & 0.009 & \textbf{3.76}  \\
Khan et al. \cite{khan13} & Yes & Known & 0.059 & 0.009 & 3.87  \\
Deeb et al. \cite{Deeb2017} & Yes & Known & 0.046 & 0.008 & 3.82  \\
Deeb et al. \cite{Deeb2017} & No & Known & 0.061 &  0.012 & 5.62  \\
Ours & No & Unknown & \textbf{0.045 } &  \textbf{0.004 } & 4.12 & \\

\hline
\end{tabular}
\end{center}
\end{table*}
\setlength{\tabcolsep}{1.4pt}

\begin{figure*}
    \centering
    \begin{minipage}{\textwidth}
    \centering
    \qquad
    Real Images
    \vspace{3mm}
    \end{minipage}
    \begin{subfigure}[b]{0.43\textwidth}
     \quad
        \includegraphics[width=\textwidth]{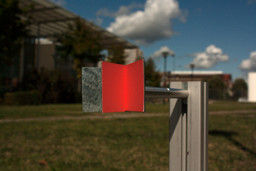}
        \label{fig:MunImg}
    \end{subfigure}
    \qquad
    ~ 
    \begin{subfigure}[b]{0.43\textwidth}
        \includegraphics[width=\textwidth]{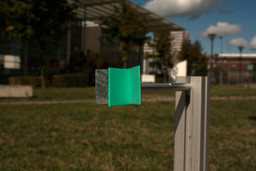}
        \label{fig:CyanImg}
    \end{subfigure}
  \\
  \begin{minipage}{\textwidth}
    \centering
    \qquad
    Surface Spectral Reflectance
    \vspace{2mm}
    {\scriptsize
    \\ {\color{blue}\rule[.5ex]{1.7em}{.6pt}} Ground Truth
    \\ {\color{red}\rule[.5ex]{.5em}{.6pt}\,\rule[.5ex]{.1em}{.6pt}\,\rule[.5ex]{.5em}{.6pt}} Estimated
    }
    
    \vspace{3mm}
    \end{minipage}
    \begin{subfigure}[b]{0.42\textwidth}
        \includegraphics[width=\textwidth]{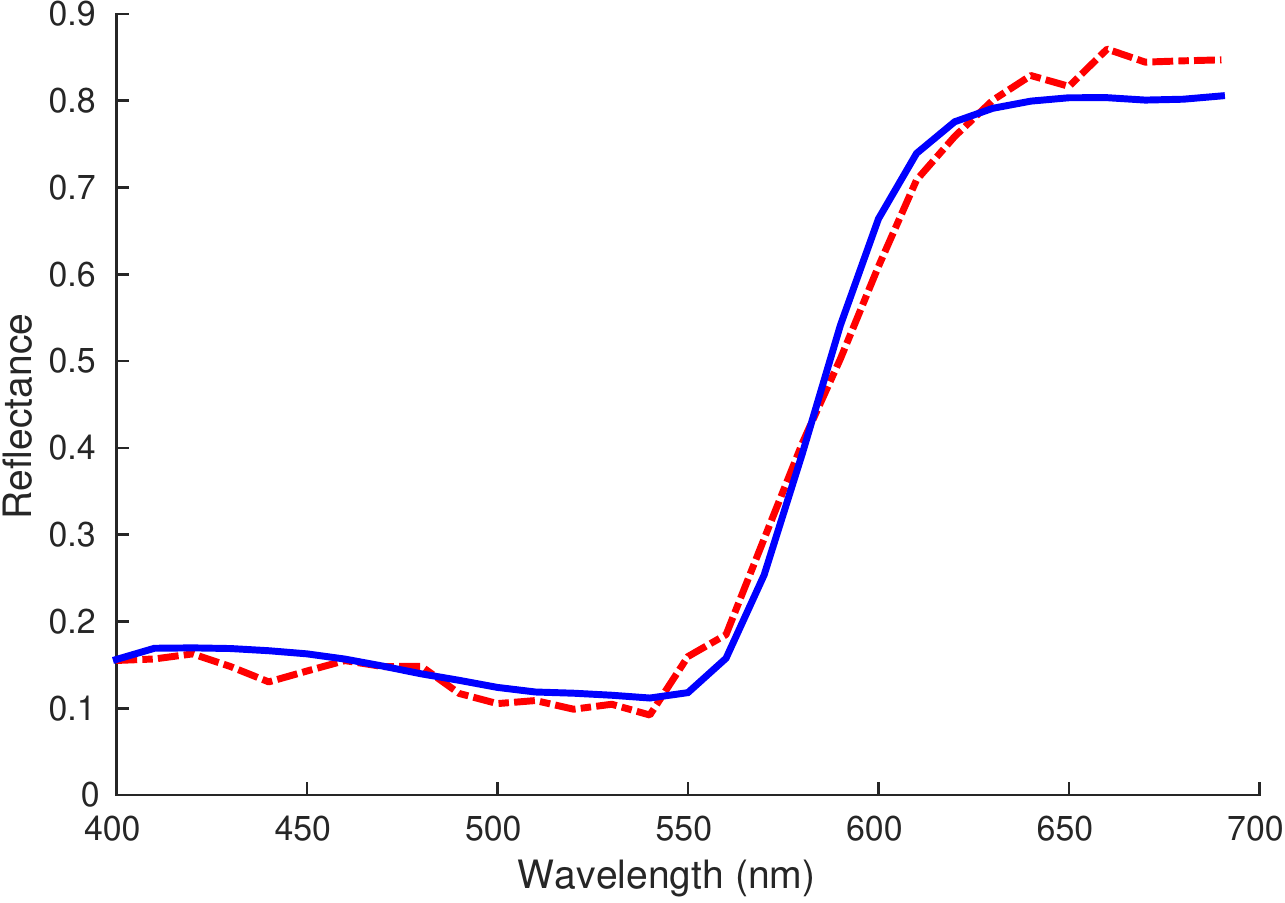}
        \label{fig:MunR}
    \end{subfigure}
    \qquad
    \begin{subfigure}[b]{0.42\textwidth}
        \includegraphics[width=\textwidth]{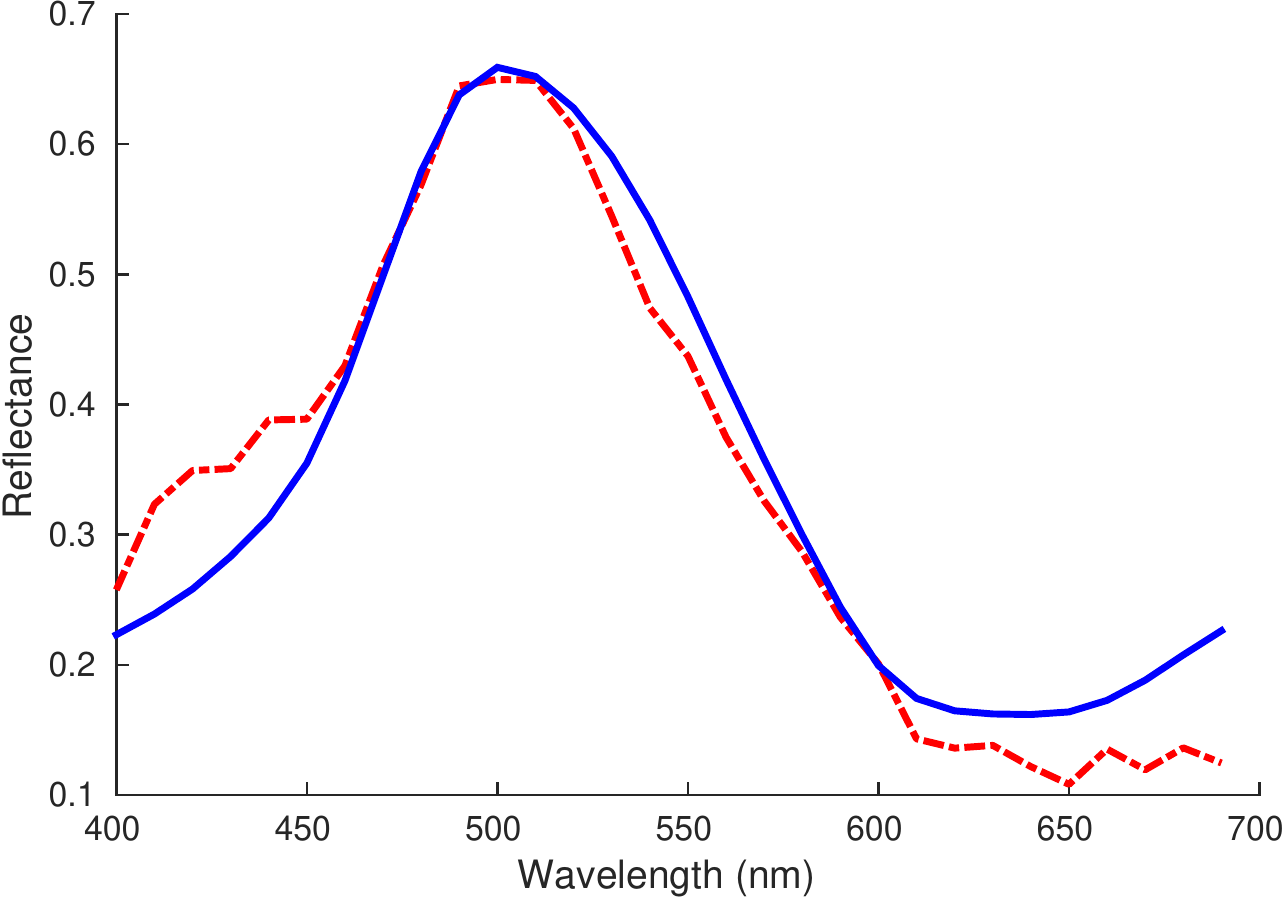}
        \label{fig:munR}
    \end{subfigure}
    \\
    \begin{minipage}{\textwidth}
    \centering
    \qquad
    Light SPD
    \vspace{2mm}
    {\scriptsize
    \\ {\color{red}\rule[.5ex]{1.7em}{.6pt}} Estimated
    \\ {\color{green}\rule[.5ex]{.5em}{.6pt}\,\rule[.5ex]{.5em}{.6pt}\,\rule[.5ex]{.5em}{.6pt}} D50
    }
    \vspace{3mm}
    
    \end{minipage}
    \begin{subfigure}[b]{0.42\textwidth}
        \includegraphics[width=\textwidth]{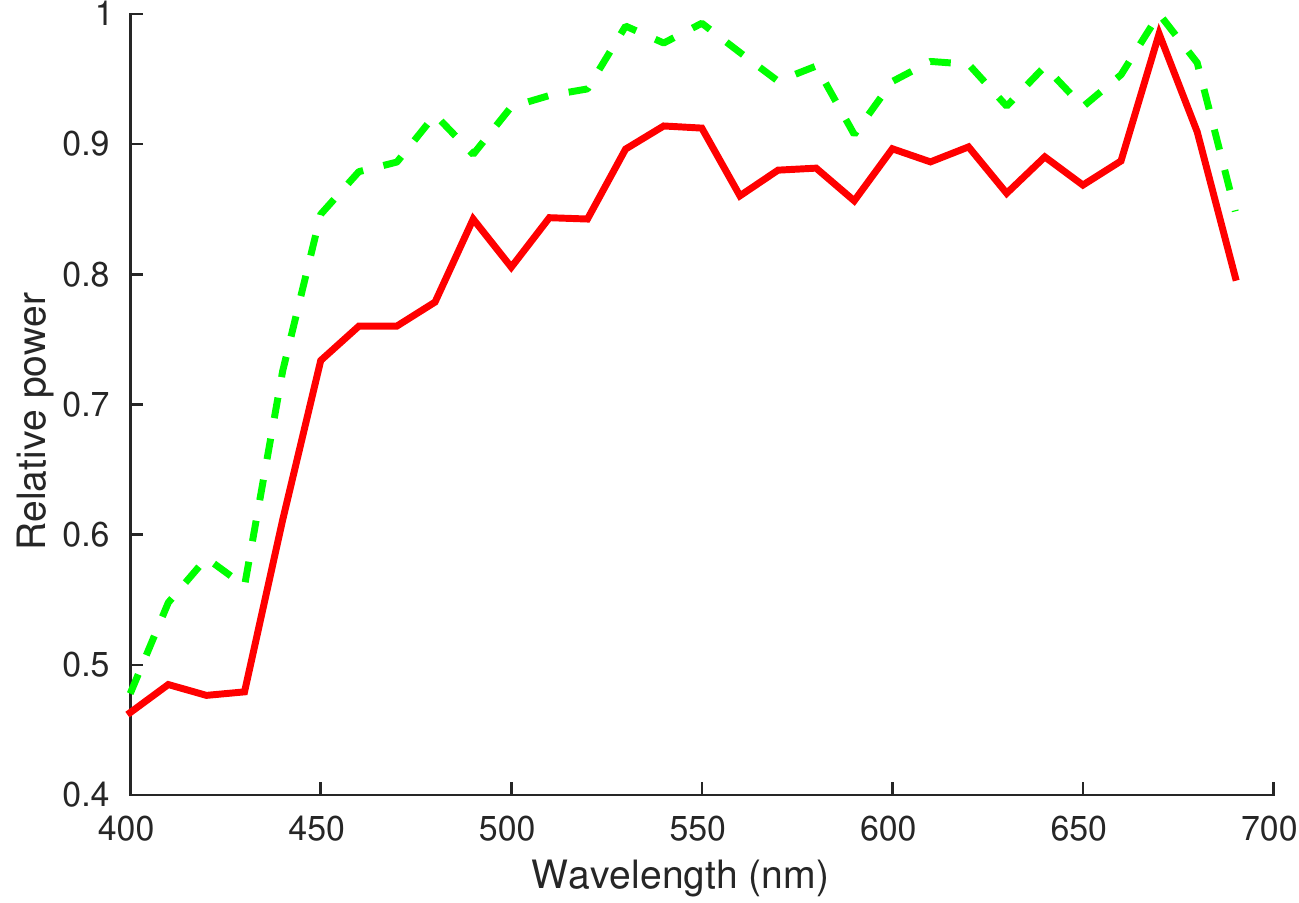}
        \label{fig:MunE}
    \end{subfigure}
    \qquad
    \begin{subfigure}[b]{0.42\textwidth}
        \includegraphics[width=\textwidth]{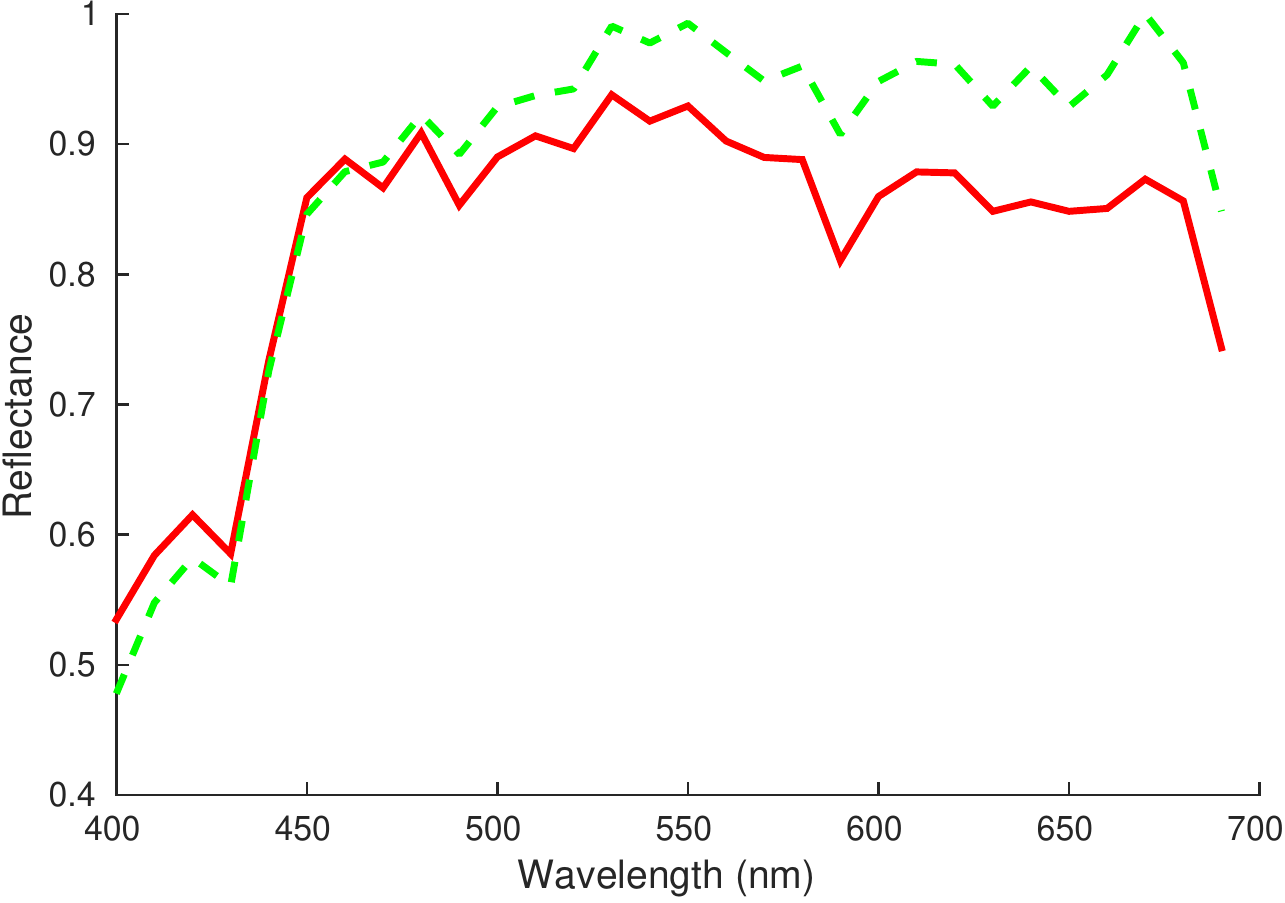}
        \label{fig:munE}
    \end{subfigure}
	\caption{Examples of estimated surface spectral reflectance and light SPD using our approach.}
    \label{fig:resReal}
\end{figure*}

Figure (\ref{fig:resReal}) shows the estimated surface spectral reflectance compared to the ground truth one using a single photo taken under daylight for Munsell red paper, and cyan textile piece. The estimated illuminant SPD is also shown next to the one of CIE D50 which was chosen in \cite{Deeb2018} as a representative of direct sunlight. 

\subsection{Generalization to other angles}
One important thing to verify is how well our approach generalizes to other geometries. One way to check this can be done by training the network on other V-shaped surfaces with different angles between the two planar surfaces. Thus, we trained the network on new datasets built using V-shaped surfaces of all Munsell patches with angles of $30^\circ$, $60^\circ$, $90^\circ$, $120^\circ$ and $150^\circ$. All the other settings are exactly the same as those used in our tests on simulated data. Table \ref{table:angResults} shows the RMSE, PD, DE00 errors regarding the spectral reflectance, as well as the RMSE error regarding the light SPD when testes on the same angle used for training. One can see that the approach generalizes well whenever the angle is small enough to give significant amount of interreflections. Our results show that both spectra are estimated with a very good accuracy for angles reaching $120^\circ$. For much bigger angles, $150^\circ$ for example, interreflections happen less leading to less important color gradients which are not enough for the network to learn the inverse solution. 

However, one possible limitation is that till now we propose to train the network on a specific geometry. Thus, and in order to provide a more concrete application of our approach, we tried to train the network on three angles together, $30^\circ$, $60^\circ$ and $90^\circ$. Table \ref{table:angResults2} shows the results in terms of RMSE, PD, and DE00 for spectral reflectance and RMSE of light SPD when the network is tested on one of the angles included in the training set, and also on a new angle which the network did not encounter before. As it can be observed, the approach generalizes very well to training on different geometries even when it is tested on a new geometry. Error values obtained when the network is tested on one of the angles used for training are very close to those obtained when the network is trained only on that specific angle (see Tables \ref{table:angResults} and \ref{table:angResults2}). On the other hand, when the network is tested on a new angle, most of the error values increase significantly. However, it is noticeable that while all the other errors increase, it is not the case for the PD one. This observation leads us to think that the network is falling in some metamerism traps for some of the patches. This is probably happening when the network is failing to recognize if the color gradients are due to high reflectance with big angle or to lower reflectance with smaller angle. Thus, the network in this case is getting the shape of the spectrum correctly but is not able to accurately identify the exact spectral reflectance. However, training on more angles with smaller steps between them would be a solution for this problem. In this case, more detailed association between the color gradients and the angles would be learned reducing as a consequence the previous case of error.

\setlength{\tabcolsep}{4pt}
\begin{table*}[ht]
	\begin{center}
		\caption{Results of training with different angles }
		\label{table:angResults} 
		\begin{tabular}{c|ccc|c}
			\hline
			& \multicolumn{3}{|c|}{Surface reflectance} & Light SPD\\
			Trained and tested angle &  RMSE &  PD   & DE00 & RMSE \\
			\hline
			$30^\circ$  & 0.0090  & 0.0014 & 0.6976 &  0.0173 \\
			$60^\circ$  & 0.0090  & 0.0014 & 0.6976 &  0.0173 \\
			$90^\circ$  & 0.0091  & 0.0014 & 0.6704 &  0.0160\\
			$120^\circ$ & 0.0093  & 0.0011 & 0.6775 &  0.0146\\
			$150^\circ$ & 0.0260  & 0.0073 & 4.1055 &   0.1050 \\			
			\hline
		\end{tabular}
	\end{center}
\end{table*}
\setlength{\tabcolsep}{1.4pt}

\setlength{\tabcolsep}{4pt}
\begin{table*}[ht]
	\begin{center}
		\caption{Results of training with different angles }
		\label{table:angResults2} 
		\begin{tabular}{cc|ccc|c}
			\hline
			&& \multicolumn{3}{c|}{Surface reflectance} & Light SPD\\
			Trained angle & Tested angle & RMSE &  PD   & DE00 & RMSE \\
			\hline
			$30^\circ$ \& $60^\circ$ \& $90^\circ$ & $60^\circ$ & 0.0099  & 0.0013 &  0.8382 & 0.0178 \\
			$30^\circ$ \& $60^\circ$ \& $90^\circ$ & $45^\circ$ & 0.0134  & 0.0014 &  1.1498
			& 0.0203 \\
			
			\hline
		\end{tabular}
	\end{center}
\end{table*}
\setlength{\tabcolsep}{1.4pt}

\section{Conclusion}\label{sec:conclusions}
In this paper, a convolutional neural network was trained in order to solve the inverse problem of surface spectral reflectance estimation form a single RGB image of interreflection under unknown lighting. Datasets were built from synthetic images simulated using infinite-bounce physics-based interreflection model. Different noise types and levels were added to the images during the training in order to better cope with noise< in camera outputs. Our experiments on simulated data showed that our approach  even under an unknown illuminant outperforms the state of the art learning-based spectral reflectance estimation approaches trained for a specific known lighting. In addition, real data results showed that our method gets a better accuracy of spectral reflectance estimation under unknown lighting than physics-based approaches under a light with a known SPD. The improvement of our method over physics-based methods could be explained by the fact that we incorporate realistic noise models in the dataset creation, whereas it is very difficult to propagate the impact of noise through physics-based methods.

However, our approach handles only direct collimated lighting. In the future, datasets can be further enhanced by using more realistic lighting conditions. This can be done for example, by modeling ambient light, shadowing and different incidence angles. In addition, the approach can be extended to take into consideration interreflection between surfaces with different spectral reflectances and under different geometrical configuration.   

\bibliography{sample}

\end{document}